\documentclass{bmvc2k}

\title{DART: Depth-as-Target Pretraining for Surgical Vision Foundation Models}

\addauthor{John Han}{john.j.han@vanderbilt.edu}{1}
\addauthor{Adam Schmidt}{}{2}
\addauthor{Muhammad Abdullah Jamal}{}{2}
\addauthor{Jie Ying Wu}{}{1}
\addauthor{Omid Mohareri}{}{2}

\addinstitution{
 Vanderbilt University\\
 2305 West End Ave\\
 Nashville, Tennessee, USA
}
\addinstitution{
 Intuitive Surgical, Inc.\\
 1020 Kifer Rd,\\
 Sunnyvale, California, USA
}

\runninghead{Han et al.}{Depth-as-target Surgical VFM}

\usepackage[utf8]{inputenc}
\usepackage[T1]{fontenc}
\usepackage{amsmath}
\usepackage{amssymb}
\usepackage{amsfonts}
\usepackage{booktabs}
\usepackage{multirow}
\usepackage{array}
\usepackage{makecell}
\usepackage{nicefrac}
\usepackage{pifont}
\usepackage{xcolor}
\usepackage{microtype}
\usepackage{url}
\usepackage{hyperref}   

\newcommand{\mymodel}{DART}

\begin{document}

\maketitle

\begin{abstract}
Vision foundation models (VFMs) are valuable in data-scarce domains such as surgery, where a single pretrained backbone can provide rich representations for many downstream tasks. Yet the dominant self-supervised pretraining paradigm uses only RGB images, leaving readily available complementary signals, such as depth maps, unused. This is a particular missed opportunity in surgery, where natural-image VFMs transfer poorly while the scene geometry is rich and informative. With strong off-the-shelf models now able to produce pseudo-labeled dense depth for any image corpus, we hypothesize that such signals can be folded into pretraining to learn better representations. We present \mymodel{}, an RGB-D pretraining recipe that builds on DINOv2 with a simple modification: a pixel-space depth reconstruction objective applied to masked iBOT patches, supervised by pseudo-labeled depth. Depth is used only during pretraining, so fine-tuning and inference remain RGB-only. We find that this pixel-level reconstruction head improves representation quality rather than disrupting it. We further show that depth, which encodes scene geometry, is more effective as a target than alternative dense signals such as Canny edges, confirming that the gains stem from depth rather than added supervision alone. Across eight surgical benchmarks spanning segmentation, depth estimation, and image-level recognition, \mymodel{} outperforms both natural-image and in-domain baselines, including a vanilla DINOv2 trained on identical data, improving dense prediction while also strengthening image-level understanding. More broadly, \mymodel{} shows that freely available geometric pseudo-labels can strengthen foundation model pretraining without extra labels or added inference cost, pointing toward stronger backbones for surgery. We will release our code and models upon acceptance.
\end{abstract}

\section{Introduction}
\label{sec:intro}

In recent years, artificial intelligence (AI) has become a driving force in novel surgical systems, powering a wide range of intraoperative systems for gesture recognition~\cite{atoum2026motion}, instrument and anatomy segmentation~\cite{endovis18}, tissue tracking~\cite{schmidt2024tracking}, automated skill assessment~\cite{liu2021towards}, and more~\cite{guni2024artificial}. Robot-assisted surgery in particular benefits from AI for higher-order capabilities such as 3D reconstruction for improved spatial awareness~\cite{wang2022neural}, natural-language interfaces between surgeon and system~\cite{perez2025surglavi,perez2026sureon}, and surgical autonomy~\cite{da2021automating,kim2024surgical,kim2025srt,acar2025monocular}. A common prerequisite to all of these tasks is robust visual understanding that simultaneously captures high-level image semantics, fine-grained semantic detail, and spatial structure. Building a model that learns such a representation is a foundational step toward intelligent surgical systems which can reduce surgeon workload, streamline surgical workflows, and improve patient outcomes~\cite{guni2024artificial}.

Visual foundation models (VFMs) have emerged as the dominant paradigm for representation learning, due to their scalability to large unlabeled video data, generalization across downstream tasks, and practicality as universal feature extractors. Recent efforts in surgical computer vision~\cite{jaspers2025scaling} therefore focus on compiling large-scale surgical video corpora and adapting established self-supervised recipes such as DINOv2~\cite{oquab2024dinov2} and MAE~\cite{he2022mae}. The resulting backbones have been adopted for tasks including phase recognition~\cite{ramesh2023dissecting,hirsch2023self}, triplet recognition~\cite{batic2024endovit,yang2025large}, and instrument segmentation~\cite{holm2023dynamic}. Because surgical data is expensive to label while requiring expert domain knowledge, a single generalist backbone that serves many downstream tasks is especially valuable.

Among these recipes, DINOv2~\cite{oquab2024dinov2} is a particularly strong backbone for dense prediction. It learns through a self-distillation objective, enforcing consistency between different views of an image under heavy data augmentations and predicting masked patches' representations, all entirely in latent space without any pixel-level reconstruction. The resulting features have been used for a variety of downstream tasks~\cite{dinov3}, making DINOv2 exactly the kind of generalist that label-scarce surgical vision calls for. However, we observed that off-the-shelf DINOv2 and DINOv3~\cite{dinov3} models do not generalize well to surgical images, with limited performance on dense pixel and image-level tasks. In fact, a MAE~\cite{he2022mae} and a DINOv2 minimally trained on surgical images tended to outperform even the released DINOv3 checkpoint on segmentation and recognition tasks. The real need, then, is not a better off-the-shelf model but a better pre-training recipe for surgical vision foundation models.


One particular approach to improve self-supervised ViT backbones, established in the MAE family, has been to incorporate additional modalities, e.g., depth maps. For instance, MultiMAE~\cite{bachmann2022multimae} and Mask3D~\cite{hou2023mask3d} both add depth map modalities to the pixel reconstruction objective for improved performance in downstream tasks, particularly in dense pixel tasks. By adding explicit geometry in the form of pseudo-labeled depth maps, these ViT backbones produce better features for semantic and spatial understanding. Two obstacles, however, make this paradigm a poor fit for surgery. First, these methods take depth as an encoder \textit{input} and therefore expect depth at inference, which is operationally restrictive in clinical pipelines where intraoperative depth sensing is rare or computationally expensive. Second, MAE-family pretraining underperforms DINOv2 on dense prediction in our surgical setting, and the RGB-D variant does not close the gap: MultiMAE pretrained on surgical images actually underperforms a single-modality MAE baseline. This suggests the path forward is not more MAE, but bringing depth supervision to DINOv2.


We present DART (Depth-as-tARgeT), a DINOv2-based framework for RGB-D pre-training in the surgical domain. We choose DINOv2 over other popular models such as I-JEPA~\cite{ijepa} precisely because the iBOT~\cite{zhou2021ibot} objective, which already supervises masked patches, offers a natural insertion point for additional pixel-space supervision that latent-only frameworks otherwise lack. In short, we add a lightweight head to reconstruct depth maps from masked iBOT patch tokens, supervised by pseudo-labeled depth, while maintaining all other loss terms and objectives as the original DINOv2. Crucially, we require no depth maps at runtime/fine-tuning, which we ensure by only adding depth as a supervisory \textit{target} rather than an input. By enforcing the model to reconstruct depth patches at masked iBOT locations, we encourage the model to learn better spatial features for both dense pixel and image-level tasks. The depth head is discarded after pre-training; at fine-tuning and inference, \mymodel{} operates purely in RGB, thus improving its practicality and deployment usability. Although our application is in surgery, our proposed method is generalizable to any specific domain. 

Across five segmentation benchmarks, one depth estimation benchmark, and two image-level surgical benchmarks, \mymodel{} simultaneously outperforms its baselines on dense prediction while maintaining strong performance on image-level surgical understanding. We further provide controlled ablations, isolating the depth objective from in-domain pre-training and characterizing design choices. Our contributions are as follows:
\begin{itemize}
    \item We propose \mymodel{}, a RGB-D self-supervised pretraining recipe that incorporates pseudo-depth as a target, requiring only RGB at fine-tuning and inference.
    \item We pre-train \mymodel{} and other self-supervised vision models on the surgical LEMON dataset~\cite{che2025lemon}. We demonstrate that \mymodel{} outperforms other models trained on the same data and their publicly released checkpoints.
    \item We also demonstrate that \textit{depth} is the preferred supervisory target in our framework. By exchanging the target with other auxiliary signals (e.g., Canny edges), we observe that depth maps generally lead to superior results than other auxiliary targets.
\end{itemize}
\section{Related Work}
\label{sec:related}

\subsection{Surgical Vision Foundation Models}
\label{sec:related-surgical}

The scarcity of large annotated datasets in surgery has motivated several efforts to develop domain-specific vision foundation models via self-supervised pretraining on unlabeled surgical video. EndoViT~\cite{batic2024endovit} pretrains a ViT with masked image modeling on a large endoscopic corpus, while EndoFM~\cite{wang2023endofm} and its long-video extension EndoFM-LV~\cite{tian2025endofmlv} adopt masked video modeling within a teacher-student framework. GSViT~\cite{schmidgall2024gsvit} instead pretrains via next-frame prediction with a focus on real-time deployment, and Hirsch et al.~\cite{hirsch2023ssl} apply Masked Siamese Networks~\cite{assran2022msn} to endoscopic video.

More recent work scales both data and supervision. SurgeNetXL~\cite{jaspers2025surgenetxl} pretrains DINO-style models on 4.7M surgical frames, and SurgFM~\cite{che2025lemon} applies DINO-based self-distillation with an augmented teacher-student scheme. A separate line explores vision-language pretraining, aligning surgical video with narration transcripts~\cite{yuan2025surgvlp,yuan2024hecvl}. Others adapt off-the-shelf models to surgical tasks; Surgical-DINO~\cite{cui2024surgicaldino}, for instance, uses LoRA-based adaptation of DINOv2 for surgical depth estimation.

Our work is most directly comparable to image-based surgical self-supervised learning (SSL) methods that pretrain a general-purpose backbone for diverse downstream tasks (EndoViT, SurgeNetXL, SurgFM). We differ in two ways: (i) we build on DINOv2 rather than MAE or DINO, retaining its strong dense prediction capabilities, and (ii) we introduce auxiliary depth supervision that leverages the geometric structure of surgical scenes.

\subsection{RGB-D Pretraining}
\label{sec:related-rgbd}

Leveraging geometric or multimodal supervision during pretraining has been studied extensively for indoor scene understanding under the masked-image-modeling paradigm. Pri3D~\cite{hou2021pri3d} uses multi-view RGB-D contrastive constraints to embed 3D priors into 2D representations, and Mask3D~\cite{hou2023mask3d} brings this into the masked-image-modeling framework by masking and reconstructing depth alongside RGB. Both establish that depth, used as a pretraining signal, transfers geometric inductive biases to RGB-only finetuning.

A related line of work treats multiple modalities jointly. MultiMAE~\cite{bachmann2022multimae} extends MAE to accept and predict RGB, depth, and semantic segmentation, and can be finetuned with any subset of these modalities. Subsequent work combines contrastive and reconstructive objectives for RGB-D~\cite{yang2023comae,jamal2025mmcmae} or scales to many modalities via tokenization~\cite{mizrahi20234m,jamal2023m3d}. Across this MAE-family of work, depth typically serves as \emph{both} encoder input and reconstruction target, so the resulting backbone expects depth at inference. However, this is a requirement that is impractical in domains without depth sensors, including most clinical pipelines.

\begin{figure}
    \centering
    \includegraphics[width=\linewidth]{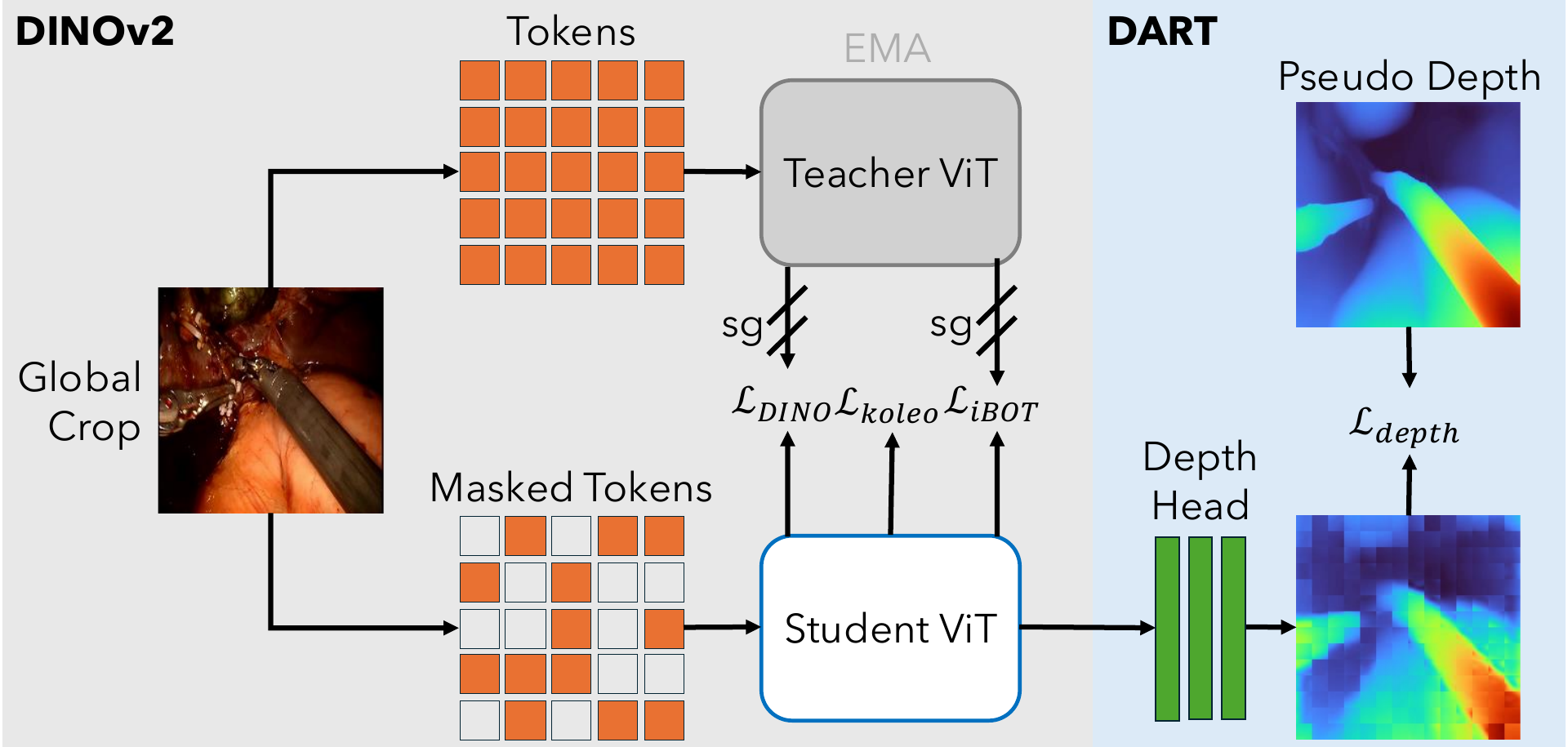}
    \caption{An overview of our method for a single global crop example. We build on the DINOv2 framework and maintain all components and loss terms. In DINOv2, global crops are 50\%-masked and enter the student backbone. The iBOT objective enforces student patch features to match the teacher's patch at the masked locations. \mymodel{} adds a lightweight head that regresses pixel-wise depth from the student's patch features. A reconstruction $\ell_1$ loss is minimized to match pseudo-labeled depth maps. Loss is computed only at masked patch locations, similar to iBOT. This simple modification lends itself to a RGB-D pre-training recipe that improves downstream tasks without requiring depth maps at fine-tuning/runtime. Local crops are processed identically to the DINOv2 recipe. ``sg'' stands for stop-gradient.}
    \label{fig:teaser}
\end{figure}

DINO-family self-distillation has seen comparatively little exploration with depth or multimodal supervision, and existing work differs from our setting. DeFM~\cite{patel2026defm} pretrains a DINOv2-style model entirely on depth images, yielding a depth encoder used on depth inputs at inference. The Omnivorous Vision Encoder~\cite{kabra2026omnivorous} adapts a \emph{frozen} DINOv2 backbone post-hoc to produce modality-agnostic features, rather than modifying the pretraining objective. To the best of our knowledge, no prior work incorporates depth as a \emph{target} within the DINOv2 pretraining objective itself, with depth never entering the encoder input.

Finally, concurrent surgical work has begun using monocular depth for self-supervised representation learning, though not within a DINOv2 framework: EndoUFM~\cite{mao2025endoufm} uses Depth Anything~\cite{yang2024depth} as a frozen prior to improve depth estimation, but does not modify the backbone's pretraining objective. In contrast, we use pseudo-depth during pretraining itself, baking geometric awareness into the encoder.

\begin{figure}
    \centering
    \includegraphics[width=0.9\linewidth]{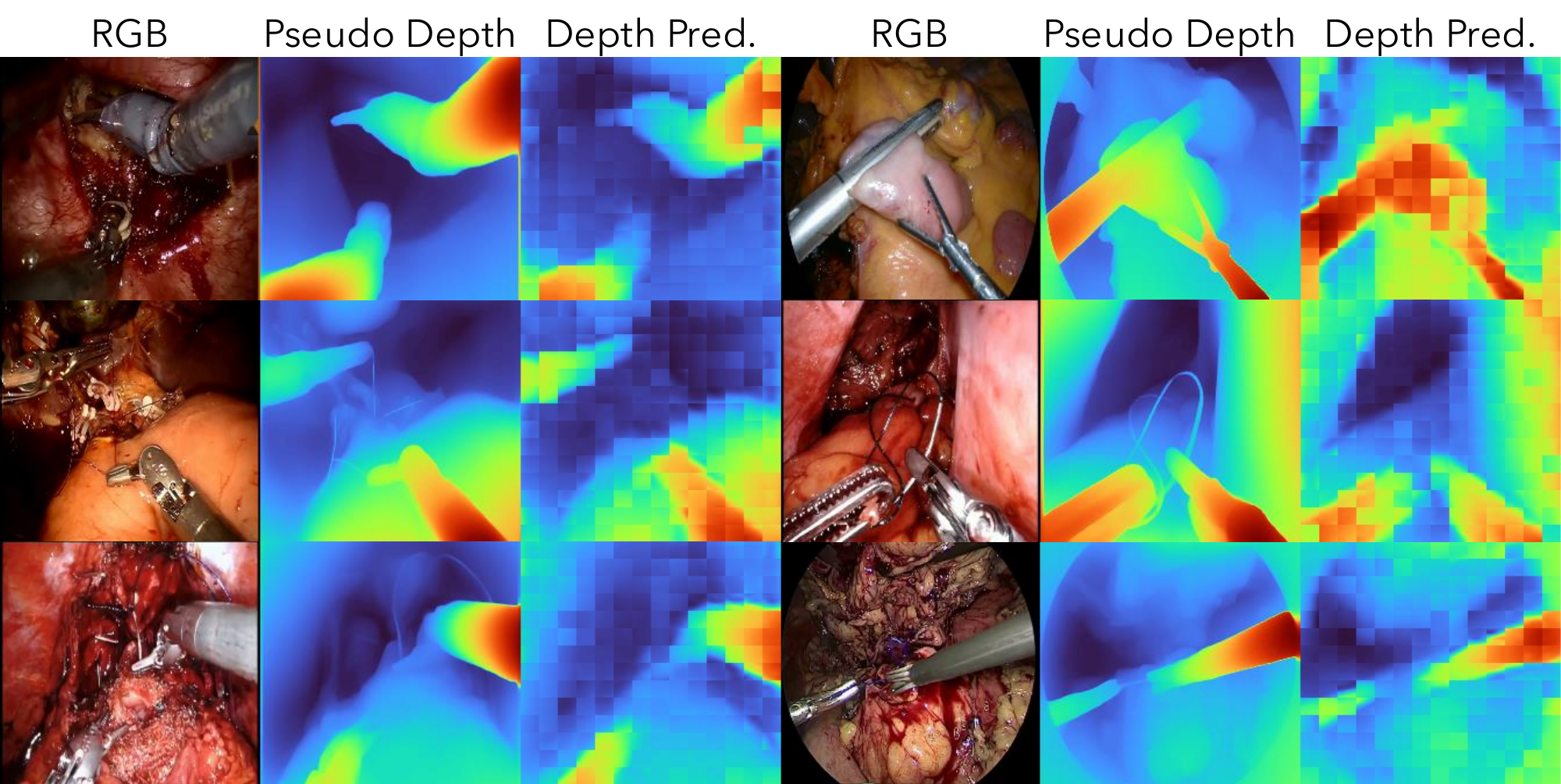}
    \caption{Samples from the RGB-D dataset and qualitative results displaying the depth reconstruction quality of \mymodel{}. Each sample is ordered as RGB, pseudo-labeled depth, and reconstructed depth of \mymodel{}. Note that the objective of our approach is not for good depth estimation but rather improving representation learning for downstream tasks.}
    \label{fig:depthreco}
\end{figure}
\section{Methods}

We propose a depth-reconstruction objective for self-supervised pretraining of vision transformers on surgical images. Our method takes RGB images as input during both pretraining and downstream finetuning; depth is used only as a supervision target during pretraining, supplied as pseudo-labels from an off-the-shelf depth predictor. The resulting encoder produces representations that improve dense surgical prediction tasks (segmentation, depth estimation) and image-level tasks (classification, recognition) without requiring depth at inference time. We build on DINOv2~\cite{oquab2024dinov2} as our base self-supervised framework, adding a lightweight depth decoder that predicts pseudo-depth from masked patch features. We first provide details about the pre-training RGB-D dataset (Sec.~\ref{sec:dataproc}), review the relevant components of DINOv2 (Sec.~\ref{sec:prelim}), describe our additions (Sec.~\ref{sec:method}), and provide implementation details (Sec.~\ref{sec:impl}). An overview of our method can be seen in Fig.~\ref{fig:teaser}. 

\subsection{Data Preprocessing}\label{sec:dataproc}
We use the recently released LEMON dataset~\cite{che2025lemon}, which consists of 938 hours of surgical video collected from YouTube. We sample each frame at 1\,Hz to generate a pre-training dataset of 3.4M images. There is no overlap between LEMON and any of the downstream datasets that we use.

Due to physical constraints of the clinical endoscope, depth sensors (e.g., LiDAR) are not currently available to collect ground-truth depth in surgical scenes. We therefore use an off-the-shelf model, Depth Anything V2-Large~\cite{yang2024depth}, to generate a relative pseudo-depth map for each image, which we standardize to zero mean and unit variance per image. These depth maps are used only during pre-training and discarded at fine-tuning and inference. We display visualizations of some RGB-D samples in Fig.~\ref{fig:depthreco}. 

\subsection{Preliminaries: DINOv2}
\label{sec:prelim}

DINOv2 trains a vision transformer by self-distillation between a student and an architecturally identical teacher, whose parameters are an exponential moving average (EMA) of the student's: $\theta_T \leftarrow m \cdot \theta_T + (1-m) \cdot \theta_S$. Three losses operate on the student's outputs. The \textit{DINO} loss produces two global and several local views of each image via random crops and color jitter, projects each view's class token through a shared head $g(\cdot)$ into a distribution over $K$ prototypes, and trains the student to match the teacher's distributions across the other views; the student processes all views while the teacher processes only the global ones. The \textit{iBOT}~\cite{zhou2021ibot} loss adds patch-level supervision: a subset $\mathcal{M}$ of the student's global-view input patches is replaced by a learned mask token, and the student's patch tokens $p^S_i$ for $i \in \mathcal{M}$ are trained, through the same head $g$, to match the teacher's tokens $p^T_i$ at those positions, where the teacher sees the unmasked image. Both are cross-entropies between the student distribution at temperature $\tau_S$ and the centered-and-sharpened teacher distribution at temperature $\tau_T$. The \textit{KoLeo} regularizer is a repulsive term on student class tokens that penalizes each sample's distance to its nearest neighbor within the batch, encouraging spread on the unit hypersphere. The full DINOv2 objective is
\begin{equation}
\mathcal{L}_\mathrm{DINOv2} = \mathcal{L}_\mathrm{DINO} + \mathcal{L}_\mathrm{iBOT} + \lambda_\mathrm{KoLeo} \mathcal{L}_\mathrm{KoLeo}.
\end{equation}
Notably, neither DINO nor iBOT requires any pixel-level reconstruction; both operate in the latent space. Despite this, DINOv2 produces useful features for semantic and spatial understanding as a result of the multi-view consistency objective. We retain this entire framework unchanged in our method.

\subsection{Depth-Aware Pretraining}
\label{sec:method}

We add a single auxiliary objective: a lightweight decoder predicts pseudo-depth at iBOT-masked patch positions from the student's patch features. We elaborate on each design choice below.

\paragraph{Pseudo-depth as supervision target.} We use the pseudo-depth maps from Sec.~\ref{sec:dataproc} purely as a supervision \textit{target}: depth never enters the encoder's input, so the student must recover geometry from RGB-derived features alone. 

\paragraph{Depth decoder.} A shallow depth head $h(\cdot)$ maps each student patch feature $p^S_i \in \mathbb{R}^D$ to a $P \times P$ depth patch (where $P$ is the encoder's patch size):
\begin{equation}
\hat{d}_i = h(p^S_i) \in \mathbb{R}^{P \times P}.
\end{equation}

We experimented with different $h(\cdot)$ designs and empirically determined that a 3-layer MLP with hidden dim 512 produced the best performance. We use the student's post-encoder, pre-projection patch features (i.e., the same features iBOT operates on, before the projection head $g$). This decoder only exists for the student network. 

\paragraph{Depth reconstruction loss.} The decoder is supervised at iBOT-masked positions only, using the same masks $\mathcal{M}$ as the iBOT loss:
\begin{equation}
\mathcal{L}_\mathrm{depth} = \frac{1}{|\mathcal{M}|} \sum_{i \in \mathcal{M}} \bigl\| \hat{d}_i - d_i \bigr\|_1
\end{equation}
where $d_i$ is the standardized pseudo-labeled depth patch at position $i$. Restricting the loss to masked positions matches iBOT's design: the encoder must produce features at masked positions sufficient to recover both semantic content (via iBOT distillation) and geometric content (via depth reconstruction).

\paragraph{Combined objective.} The final pretraining loss is
\begin{equation}
\mathcal{L} = \mathcal{L}_\mathrm{DINO} + \mathcal{L}_\mathrm{iBOT} + \lambda_\mathrm{KoLeo} \mathcal{L}_\mathrm{KoLeo} + \lambda_\mathrm{depth} \mathcal{L}_\mathrm{depth}
\end{equation}
where $\lambda_\mathrm{depth}$ is a single hyperparameter governing the strength of depth supervision relative to DINOv2's existing losses. We choose $\lambda_\mathrm{depth} = 0.1$, as it led to the best results and report sweeps over $\lambda_\mathrm{depth}$ values in section~\ref{sec:results:sweeps}. Visualizations of depth reconstructions can be seen in Fig.~\ref{fig:depthreco}. 

\subsection{Implementation Details}
\label{sec:impl}

\paragraph{Pretraining.} We pretrain ViT-B/16~\cite{dosovitskiy2021image} \textit{from scratch} as our backbone, matching DINOv2's reference configuration, and follow DINOv2 hyperparameters except where noted. We train at a global crop resolution of $224\times224$ (yielding $14\times14$ patch tokens) with 8 local crops at $96\times96$; iBOT masks 50\% of the global-view patches. We optimize with AdamW using a 1.5e-3 base learning rate under a cosine schedule with 62,500 warmup iterations, weight decay 0.04, and teacher EMA momentum 0.996. The KoLeo weight is $\lambda_\mathrm{KoLeo}=0.1$ and our depth weight is $\lambda_\mathrm{depth}=0.1$. The depth decoder $h(\cdot)$ is a 3-layer MLP with hidden dimension 512 and GELU activations, regressing each masked patch feature to a $16\times16$ depth patch under an $\ell_1$ loss. Due to computational constraints, we train \mymodel{} for 187{,}500 iterations at batch size 1024, which we use as the common training budget across our in-domain baselines for fair comparison. Pseudo-depth targets are precomputed once with Depth Anything V2-Large~\cite{yang2024depth} prior to training.

\paragraph{Fine-tuning.} For all downstream tasks we freeze the backbone and train only the task-specific head, so depth supervision affects evaluation solely through the learned features. Fine-tuning implementation details are given in the appendices. 

\paragraph{Compute.} Pretraining runs on 8$\times$H200 GPUs; fine-tuning experiments use A100s.
\section{Results}
\subsection{Experimental Design}
We evaluate \mymodel{} on a variety of surgical downstream tasks spanning dense pixel prediction (segmentation and depth estimation) and image-level recognition. Dataset details are provided in the appendix. In all experiments we freeze the backbone and train only the task-specific heads, so every reported gain is attributable to the learned features alone. We note that all downstream tasks are out-of-domain with respect to the pre-training corpus.

\textbf{Baselines.} We compare against two families of models. First, we evaluate publicly available VFMs: DINOv2~\cite{oquab2024dinov2}, DINOv3~\cite{dinov3}, MAE (IN-1k)~\cite{he2022mae}, and MultiMAE (IN-1k)~\cite{bachmann2022multimae}. The DINO models were distilled from a ViT-g and a 7B-parameter model pre-trained on 142M and 1.7B images, respectively, while the ImageNet-1k models were trained from scratch for 1600 epochs on 1.3M natural images. Additionally, we also compare with LEMON-FM~\cite{che2025lemon}, which was trained with the DINO framework with a ConvNeXt-L backbone on the LEMON dataset. 

Second, we train each of these models on the LEMON dataset for a controlled in-domain comparison. Due to computational constraints, MAE and MultiMAE are trained for 400 epochs. The original MultiMAE was pre-trained with three modalities (RGB, depth, and semantic maps); we use only RGB and depth to match our setting. Finally, and most importantly, we train a vanilla DINOv2 on the same data with identical training parameters to our method, isolating the effect of the depth objective. Unless specified otherwise, we use ViT-B for all models and fine-tune at $224\times224$ resolution. Implementation details for pre-training and fine-tuning these models are in the appendix.

\subsection{Dense Pixel Tasks}
We first report the performance of \mymodel{} against its baselines on dense pixel tasks: segmentation and monocular depth estimation. We additionally report ViT-S experiments in the appendix, showing that the gains are not specific to a single model scale.

\textbf{Segmentation.} We consider two setups following previous work~\cite{oquab2024dinov2}. The first is a linear (lin.) head, which probes the class separability of the features with a lightweight linear operation on the final layer's features: the low-resolution patch features are projected to per-class logits, which are then bilinearly upsampled to the full-resolution map. The second is a multi-scale (+ms.) head, where we concatenate patch features along the embedding dimension from the standard probing layers [2,5,8,11] and train a linear layer on top. Using intermediate features alongside final features provides additional context for the task and generally boosts performance. All experiments share the same training parameters and use a cross-entropy loss. We report mean Intersection over Union (mIoU), averaged over three seeds; due to space constraints, per-cell standard deviations are reported in the appendix.

We evaluate on five established surgical segmentation datasets: EndoVis18~\cite{endovis18}, CholecInstanceSeg~\cite{alabi2025cholecinstanceseg}, SAR-RARP50~\cite{sarrarp50}, CholecSeg8k~\cite{hong2020cholecseg8k}, and PhaKIR~\cite{phakir}. We use the standard train-test splits provided in the original datasets. Because PhaKIR does not provide a public test set, we use Video\_13 as the test set and exclude it from training. For these experiments we also include MultiMAE RGB-D, a variant that additionally receives depth as input at fine-tuning, since the original authors report that providing extra modalities at fine-tuning improves downstream performance~\cite{bachmann2022multimae}. These results are shown in Table~\ref{tab:semseg}.

\begin{table}
  \centering
  \setlength{\tabcolsep}{2.5pt}      
  \footnotesize                       
  \newcommand{\dimrow}[1]{\textcolor{gray}{#1}}
  \newcommand{\second}[1]{\underline{#1}}
  \begin{tabular}{@{}ll c cc cc cc cc cc@{}}
    \toprule
    & & & \multicolumn{2}{c}{EndoVis18} & \multicolumn{2}{c}{CholecISeg} & \multicolumn{2}{c}{SARRARP50} & \multicolumn{2}{c}{CholecSeg8k} & \multicolumn{2}{c}{PhaKIR} \\ 
    \cmidrule(lr){4-5} \cmidrule(lr){6-7} \cmidrule(lr){8-9} \cmidrule(lr){10-11} \cmidrule(lr){12-13}
    Data & Method & Depth & lin. & +ms. & lin. & +ms. & lin. & +ms. & lin. & +ms. & lin. & +ms. \\
    \midrule
    \multirow{4}{*}{Natural}
    & DINOv2           & --    & 28.8 & 31.8 & 48.3 & 52.6 & 53.5 & 62.0 & 36.7 & 38.5 & 10.3 & 10.4 \\
    & DINOv3           & --    & 30.1 & 32.5 & 53.5 & 55.2 & 58.0 & 64.0 & \second{50.0} & 43.7 & 11.6 & 10.6 \\
    & MAE      & --    & 26.0 & 28.0 & 37.1 & 47.1 & 52.7 & 55.1 & 37.9 & 38.9 &  7.6 &  8.8 \\
    & MultiMAE & PT    & 26.6 & 27.0 & 40.7 & 48.7 & 51.2 & 56.1 & 36.0 & 36.7 &  8.3 & 10.1 \\
    \midrule
    \multirow{6}{*}{LEMON}
    & LEMON-FM         & --    & 25.7 & 31.2 & 40.6 & 51.0 & 52.0 & 66.6 & 33.9 & 42.4 & 8.7 & 11.0 \\
    & MAE              & --    & 31.9 & 32.6 & 41.1 & 56.2 & \second{65.7} & \second{67.1} & 44.3 & 46.0 & 12.0  & 13.2 \\
    & MultiMAE         & PT    & 30.8 & 32.6 & \second{59.5} & 55.1 & 62.7 & 66.0 & 43.4 & 43.1 & 11.7 & 12.2 \\
    & DINOv2           & --    & \second{34.1} & \second{34.9} & 58.2 & \second{59.7} & 63.9 & 64.4 & 49.7 & \second{50.5} & \second{14.1} & \second{15.7} \\
    & \textbf{Ours}    & PT    & \bf 34.9 & \bf 36.8 & \bf 66.2 & \bf 68.9 & \bf 68.7 & \bf 70.2 & \bf 52.4 & \bf 50.5 & \bf 14.5 & \bf 16.2 \\
    & \dimrow{MultiMAE} & \dimrow{PT+FT} & \dimrow{29.4} & \dimrow{28.8} & \dimrow{54.7} & \dimrow{49.9} & \dimrow{70.4} & \dimrow{71.8} & \dimrow{44.3} & \dimrow{45.2} & \dimrow{11.0} & \dimrow{11.9} \\
    \bottomrule
  \end{tabular}
  \caption{\textbf{Semantic segmentation with frozen features}, linear (lin.) and
  multiscale (+ms). \emph{Depth} marks whether depth is used in pre-training (PT)
  and/or fine-tuning (FT). \textbf{Bold} is best and \underline{underline} is second best among the RGB-only fine-tuning settings; the RGB-D fine-tuning row (gray) is excluded from the ranking. Mean over 3 seeds; per-cell std reported in the appendix. CholecISeg = CholecInstanceSeg. Natural images refer to LVD for DINO models and IN-1k for MAE models. We report the mean over 3 seeds; standard deviations are in the appendices due to space constraints.}
  \label{tab:semseg}
\end{table}

\mymodel{} shows notable gains over the RGB-only DINOv2 baseline trained on identical data, demonstrating that adding depth supervision during pre-training reliably improves downstream performance. For instance, our method outperforms the DINOv2 baseline by +9.2 mIoU on CholecInstanceSeg (+ms.). In some settings the gains are modest or on par, such as CholecSeg8k multiscale and PhaKIR linear. Our method also surpasses LEMON-FM~\cite{che2025lemon}, a strong in-domain surgical foundation model, on dense prediction. This is notable given that LEMON-FM is a larger ConvNeXt-L backbone with over twice as many parameters.

More generally, publicly available VFM backbones are not competitive on surgical segmentation. MAE and MultiMAE trained for 1600 epochs on natural images do not outperform their in-domain counterparts trained for only 400 epochs on surgical images, underscoring a significant gap between the natural and surgical domains.

Surprisingly, MAE is superior to MultiMAE across most benchmarks when fine-tuning with RGB only, despite MultiMAE having been pre-trained with RGB-D; for instance, MAE reaches 31.9 mIoU on EndoVis18 linear, whereas MultiMAE reaches 30.8. This suggests that masked image modeling is a limited paradigm for RGB-D learning in the surgical domain. In some cases, however, MultiMAE is comparable to or even outperforms MAE (59.5 vs 41.1 mIoU on CholecInstanceSeg linear). MultiMAE can also be fine-tuned with depth as input (RGB-D segmentation), shown in the last row of Table~\ref{tab:semseg}, and we find that performance actually \textit{degrades} with the depth modality on most benchmarks. While we do not test this directly, we speculate that this unexpected discrepancy stems from two factors: (1) MAE-family models are far more competitive under end-to-end fine-tuning than the frozen setting evaluated here, and (2) adding depth as input introduces an additional domain shift between pre-training and fine-tuning. Notably, \mymodel{} outperforms RGB-D MultiMAE on 8 of 10 benchmark settings while using only RGB at fine-tuning.

\begin{figure}
    \centering
    \includegraphics[width=\linewidth]{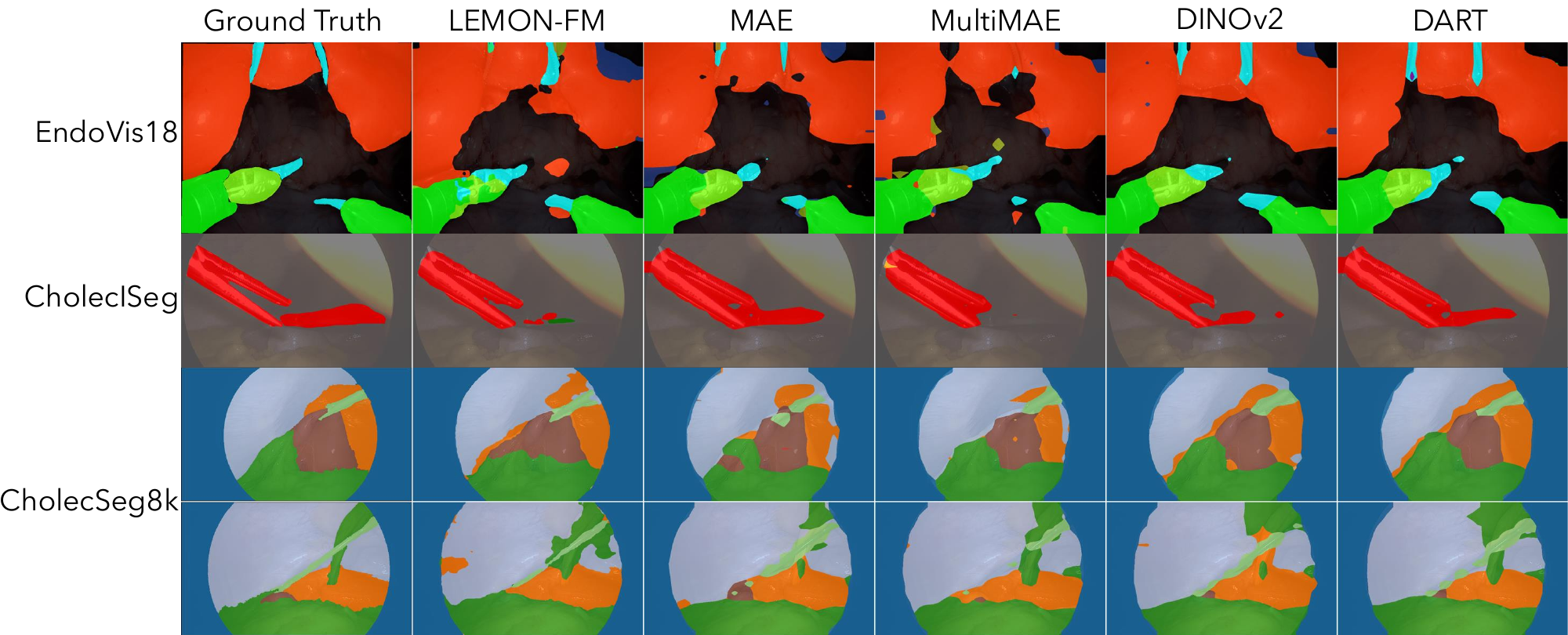}
    \caption{\textbf{Qualitative results displaying segmentation results} on EndoVis18, CholecInstanceSeg, and CholecSeg8k under the multiscale setting. We only display results for models pretrained on the surgical data. We refer the reader to the appendices for more qualitative results.}
    \label{fig:seg_qual}
\end{figure}


We visualize selected qualitative results in Fig.~\ref{fig:seg_qual} for 3 segmentation datasets over the 5 LEMON-trained models. In general, \mymodel{} produces cripser instrument boundaries with less false positive regions. Other models tend to produce noisy semantic maps with small incorrectly classified regions compared. In particular, even DINOv2 produces noisy semantic maps, despite the fact that the only difference between \mymodel{} and DINOv2 is the depth-as-target supervision. 

\textbf{Depth Estimation.} We again consider two setups: a linear head that regresses a single channel from the final layer's patch features, and the DPT~\cite{dpt} head (where we take features from encoder layers [2,5,8,11]). We train both heads with an $\ell_1$ loss and do not use the class token.

Because ground-truth depth is difficult to acquire in surgical scenes, labeled surgical depth datasets are scarce; most public datasets either consist of simulated/phantom data~\cite{c3vdv2,endoslam} or rely on off-the-shelf stereo matching~\cite{recasens2021endo}. For our evaluation, we use the corrected SCARED-C dataset~\cite{scaredc}, which provides ex-vivo porcine scenes with sparse depth labels acquired via structured light. We perform relative depth estimation by predicting normalized depth maps and report $\delta_1$ in Table~\ref{tab:depth}.

\begin{table}[t]
\centering

\setlength{\tabcolsep}{8pt}
\small
\newcommand{\second}[1]{\underline{#1}}
\begin{tabular}{llccc}
\toprule
Pretrain. data & Method & Depth & Linear & DPT \\
\midrule
\multirow{5}{*}{LEMON}
    & LEMON-FM         & --    & 0.485\tiny{$\pm$0.005} & 0.608\tiny{$\pm$0.006} \\
& MAE              & --    & 0.512\tiny{$\pm$0.003} & 0.609\tiny{$\pm$0.026} \\
& MultiMAE         & PT    & 0.498\tiny{$\pm$0.004} & 0.605\tiny{$\pm$0.007} \\
& DINOv2           & --    & \second{0.542}\tiny{$\pm$0.003} & 0.626\tiny{$\pm$0.009} \\
& \textbf{Ours}    & PT    & \textbf{0.557}\tiny{$\pm$0.008} & \textbf{0.653}\tiny{$\pm$0.009} \\
\midrule
\multirow{4}{*}{\textcolor{gray}{Natural}}
& \textcolor{gray}{DINOv2} & \textcolor{gray}{--} & \textcolor{gray}{0.506\tiny{$\pm$0.011}} & \textcolor{gray}{0.613\tiny{$\pm$0.037}} \\
& \textcolor{gray}{DINOv3} & \textcolor{gray}{--} & \textcolor{gray}{0.500\tiny{$\pm$0.011}} & \textcolor{gray}{0.623\tiny{$\pm$0.009}} \\
& \textcolor{gray}{MAE} & \textcolor{gray}{--} & \textcolor{gray}{0.499\tiny{$\pm$0.006}} & \textcolor{gray}{0.598\tiny{$\pm$0.003}} \\
& \textcolor{gray}{MultiMAE} & \textcolor{gray}{PT} & \textcolor{gray}{0.487\tiny{$\pm$0.004}} & \textcolor{gray}{\second{0.630}\tiny{$\pm$0.005}} \\
\bottomrule
\end{tabular}
\caption{\textbf{Depth estimation on SCARED-C ($\delta_1 \uparrow$).} Mean$\pm$std over 3 seeds. \textbf{Bold} is best and \underline{underline} is second best across all rows. We exclude MultiMAE RGB-D since it would be a trivial task to regress depth from RGB-D. \emph{Depth} marks whether depth is used in pre-training (PT). Natural images refer to LVD and IN-1k for DINO and MAE models respectively.}
\label{tab:depth}
\end{table}

We observe a trend similar to the segmentation results: our method outperforms its baselines under both the linear and DPT heads. Here, however, publicly available VFMs are far more competitive, with DINOv3-DPT rivaling many models trained on surgical images. We attribute this to SCARED-C consisting of ex-vivo porcine scenes rather than in-vivo surgical images, which likely widens the gap between pre-training and fine-tuning data and limits models pre-trained on surgical video. Nonetheless, \mymodel{} still achieves the highest accuracy, with a $\delta_1$ of 0.653 versus 0.630 for the next-best model. As before, simply adding depth during pre-training does not by itself help depth estimation: MultiMAE lags behind MAE regardless of training data. In contrast, our depth-as-target objective improves over the DINOv2 baseline.

\begin{figure}
    \centering
    \includegraphics[width=0.9\linewidth]{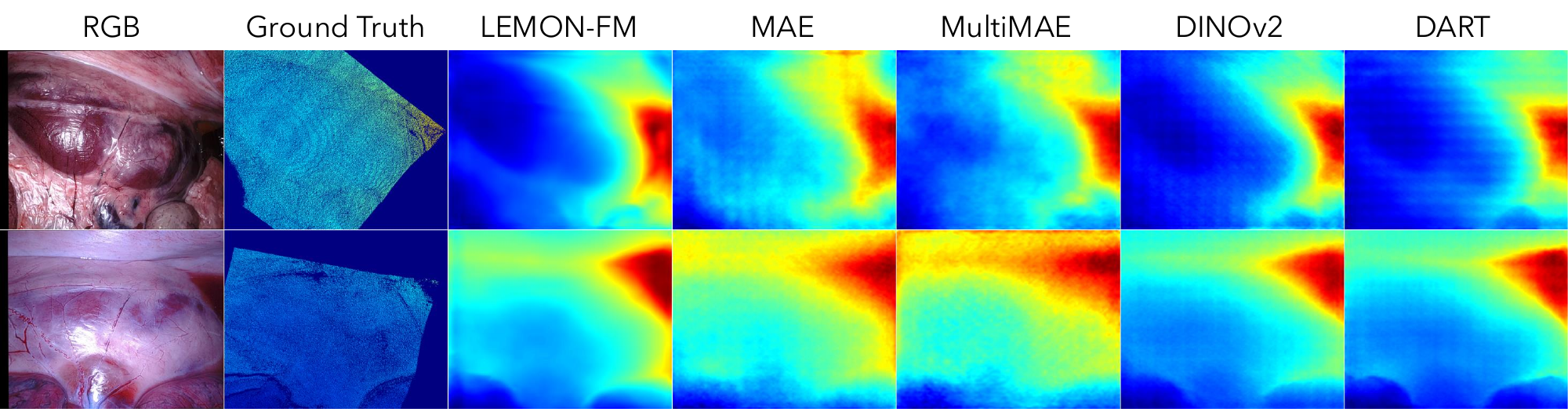}
    \caption{\textbf{Qualitative results displaying depth estimation} results on the SCARED-C dataset under the DPT setting. We only display results for models pretrained on the surgical data. We refer the reader to the appendices for more qualitative results.}
    \label{fig:depth_qual}
\end{figure}

We visualize some qualitative results on the SCARED-C dataset across the five surgical models in Fig.~\ref{fig:depth_qual}. LEMON-FM and DINOv2 tend to produce smoother depth maps. However, \mymodel{} has finer delineations and boundaries, such as the small protruding regions in the bottom of both images. MultiMAE seems to display the worst results, despite the fact that it was trained with the depth modality during pre-training. 

\subsection{Image-level Tasks}
We report two tasks: surgical phase recognition with Cholec80~\cite{cholec80} and triplet recognition with CholecT50~\cite{nwoye2022data}. Results are shown in Table~\ref{tab:image_level}.

\begin{table*}[t]
\centering

\setlength{\tabcolsep}{4pt}
\small
\newcommand{\second}[1]{\underline{#1}}
\begin{tabular}{ll@{\hskip 4pt}c cccccc}
\toprule
& & Cholec80 & \multicolumn{6}{c}{CholecT50 Triplet Recognition} \\
\cmidrule(lr){3-3} \cmidrule(lr){4-9}
Pretrain. data & Method & F1 Score & AP$_i$ & AP$_v$ & AP$_t$ & AP$_{it}$ & AP$_{iv}$ & AP$_{ivt}$ \\
\midrule
\multirow{5}{*}{LEMON}
& MAE              & 45.0\tiny{$\pm$0.4} & 36.4\tiny{$\pm$0.1} & 23.9\tiny{$\pm$0.2} & 15.8\tiny{$\pm$0.5} & 5.9\tiny{$\pm$0.1}  & 9.6\tiny{$\pm$0.2}  & 3.6\tiny{$\pm$0.1} \\
& MultiMAE         & 43.9\tiny{$\pm$0.3} & 37.8\tiny{$\pm$0.3} & 23.7\tiny{$\pm$0.2} & 15.7\tiny{$\pm$0.4} & 5.2\tiny{$\pm$0.2}  & 9.5\tiny{$\pm$0.1}  & 3.5\tiny{$\pm$0.3} \\
& DINOv2           & \second{70.8}\tiny{$\pm$0.1} & 58.8\tiny{$\pm$0.5} & 34.6\tiny{$\pm$0.2} & 25.9\tiny{$\pm$0.1} & 10.4\tiny{$\pm$0.5} & 15.4\tiny{$\pm$0.1} & 6.4\tiny{$\pm$0.1} \\
& Ours           & \textbf{71.5}\tiny{$\pm$0.1} & \second{62.1}\tiny{$\pm$0.2} & \second{37.0}\tiny{$\pm$0.1} & \second{28.2}\tiny{$\pm$0.2} & \second{11.6}\tiny{$\pm$0.1} & \second{16.2}\tiny{$\pm$0.5} & \second{6.9}\tiny{$\pm$0.1} \\
& LEMON-FM           & 67.1\tiny{$\pm$0.1} & \textbf{73.6}\tiny{$\pm$0.4} & \textbf{44.6}\tiny{$\pm$0.2} & \textbf{30.0}\tiny{$\pm$0.1} & \textbf{13.3}\tiny{$\pm$0.4} & \textbf{20.0}\tiny{$\pm$0.1} & \textbf{7.9}\tiny{$\pm$0.3} \\
\midrule
\multirow{4}{*}{\textcolor{gray}{Natural}}
& \textcolor{gray}{DINOv2}   & \textcolor{gray}{60.7\tiny{$\pm$0.3}} & \textcolor{gray}{55.9\tiny{$\pm$0.2}} & \textcolor{gray}{34.8\tiny{$\pm$0.2}} & \textcolor{gray}{26.4\tiny{$\pm$0.2}} & \textcolor{gray}{10.3\tiny{$\pm$0.1}} & \textcolor{gray}{15.1\tiny{$\pm$0.0}} & \textcolor{gray}{6.2\tiny{$\pm$0.0}} \\
& \textcolor{gray}{DINOv3}   & \textcolor{gray}{56.6\tiny{$\pm$0.0}} & \textcolor{gray}{46.7\tiny{$\pm$0.2}} & \textcolor{gray}{29.0\tiny{$\pm$0.1}} & \textcolor{gray}{20.6\tiny{$\pm$0.2}} & \textcolor{gray}{8.0\tiny{$\pm$0.0}}  & \textcolor{gray}{13.0\tiny{$\pm$0.0}} & \textcolor{gray}{4.8\tiny{$\pm$0.0}} \\
& \textcolor{gray}{MAE}      & \textcolor{gray}{25.9\tiny{$\pm$0.4}} & \textcolor{gray}{33.1\tiny{$\pm$0.2}} & \textcolor{gray}{19.6\tiny{$\pm$0.5}} & \textcolor{gray}{11.9\tiny{$\pm$0.1}} & \textcolor{gray}{4.2\tiny{$\pm$0.1}}  & \textcolor{gray}{8.0\tiny{$\pm$0.2}}  & \textcolor{gray}{2.5\tiny{$\pm$0.4}} \\
& \textcolor{gray}{MultiMAE} & \textcolor{gray}{44.2\tiny{$\pm$0.3}} & \textcolor{gray}{39.7\tiny{$\pm$0.2}} & \textcolor{gray}{25.3\tiny{$\pm$0.3}} & \textcolor{gray}{17.6\tiny{$\pm$0.1}} & \textcolor{gray}{5.6\tiny{$\pm$0.1}}  & \textcolor{gray}{10.2\tiny{$\pm$0.1}} & \textcolor{gray}{3.5\tiny{$\pm$0.0}} \\
\bottomrule
\end{tabular}
\caption{\textbf{Image-level surgical understanding.} Cholec80 phase recognition F1 Score (\%) and CholecT50 triplet recognition (AP, \%). Mean$\pm$std over 3 seeds. \textbf{Bold} = best and \underline{underline} = second best per column, across all rows. Natural image datasets are LVD for DINO models and IN-1k for MAE models.}
\label{tab:image_level}
\end{table*}

\textbf{Cholec80.} Surgical phase recognition is the task of classifying which stage of the surgery an image depicts (e.g., preparation, dissection, cleaning). We treat this as a simple classification task: a linear layer over the class token produces the logits, trained with a cross-entropy loss, and we report F1 score. Our method again shows significant gains over its baselines. While depth is most intuitively useful for dense pixel tasks, we find it also benefits image-level understanding and classification. Specifically, our method outperforms the MAE models, the DINOv2 baseline (+0.7 F1), and the DINO-based LEMON-FM (+4.4 F1).

\textbf{CholecT50.} Surgical triplet recognition is the task of classifying the $\langle$instrument, action, target$\rangle$ tuple in a given image. Each frame may contain zero, one, or multiple triplets, making this a multi-label classification problem. We found that concatenating the cls token and the mean of the patch features worked better than using the class token alone, and that three independent linear heads (one per label) trained with a multi-label binary cross-entropy loss outperformed a single linear layer. For metrics, we use the \verb|ivtmetrics| library and report average precision following previous work~\cite{nwoye2022data}.

Triplet recognition diverges from the trends above. While our method beats most baselines, the DINO-based LEMON-FM outperforms it by a notable margin (7.9 vs 6.9 $AP_{ivt}$). We note that LEMON-FM is a ConvNeXt-L backbone with more than twice the parameters, and that it was trained with a temporally augmented distillation strategy that draws positive views from similar procedures in other videos. This improves its invariance to minor instrument and tissue motion and to appearance changes across patients, benefiting fine-grained tasks like triplet recognition. Even so, \mymodel{} improves over all baselines of comparable model size despite not utilizing a domain-specific image sampling strategy.

\subsection{Auxiliary Targets}
Our results suggest that depth supervision during pre-training provides significant downstream gains. Intuitively, encoding the 3D structure of the scene should yield better semantic and image-level understanding. But this raises a natural question: do the gains come from regularizing with any auxiliary image, or from depth specifically?

To answer this, we swap the depth maps for other spatial targets during pre-training: Canny edges and blurred grayscale images. Both alternatives contain information that could plausibly aid segmentation. For computational efficiency, we train ViT-S variants of the DINOv2 baseline and our method, each with a different supervisory target, and evaluate on the five segmentation tasks in Table~\ref{tab:aux_signal}.

\begin{table}[t]
  \centering
  \setlength{\tabcolsep}{4pt}
  \footnotesize
  \newcommand{\second}[1]{\underline{#1}}
  \begin{tabular}{@{}l cc cc cc cc cc@{}}
    \toprule
    & \multicolumn{2}{c}{EndoVis18} & \multicolumn{2}{c}{CholecISeg} & \multicolumn{2}{c}{SARRARP50} & \multicolumn{2}{c}{CholecSeg8k} & \multicolumn{2}{c}{PhaKIR} \\
    \cmidrule(lr){2-3} \cmidrule(lr){4-5} \cmidrule(lr){6-7} \cmidrule(lr){8-9} \cmidrule(lr){10-11}
    Aux.\ signal & lin. & +ms. & lin. & +ms. & lin. & +ms. & lin. & +ms. & lin. & +ms. \\
    \midrule
    None (DINOv2)     & \second{30.4} & \second{30.9} & \second{48.5} & \bf 50.9 & \second{57.4} & 54.6 & \second{39.9} & 42.7 & 10.3 &  \second{9.5} \\
    Canny edges       & 29.1 & \second{30.9} & \bf 48.6 & 49.9 & 56.3 & \second{55.3} & 38.5 & \second{43.7} & \second{10.6} &  8.7 \\
    Blurred grayscale & 28.2 & 29.2 & 45.4 & \second{50.0} & 55.9 & 53.8 & 38.4 & \bf 44.2 & 10.4 &  9.3 \\
    Depth (ours)      & \bf 31.4 & \bf 31.2 & \second{48.5} & 47.4 & \bf 57.8 & \bf 55.6 & \bf 41.0 & 42.7 & \bf 10.7 & \bf 10.7 \\
    \bottomrule
  \end{tabular}
  \caption{\textbf{Auxiliary reconstruction signal ablation} (ViT-S; linear (lin.) and multiscale (+ms.) mIoU, mean over 3 seeds). \textbf{Bold} = best and \underline{underline} = second best per column. We report mIoU.
  CholecISeg = CholecInstanceSeg.}
  \label{tab:aux_signal}
\end{table}

Across 7 of the 10 segmentation settings, depth is the most useful target in our framework, outperforming Canny or grayscale reconstruction by up to 2.5 mIoU points. There are a few outliers: plain DINOv2 is best on CholecInstanceSeg multiscale, while Canny edges leads on the same dataset under linear probing. Overall, depth is the most consistent reconstruction target, yielding the greatest improvement over to the DINOv2 baseline. This indicates that the improvements in \mymodel{} are not simply a product of more data; rather, asking the model to predict the distance from the camera to each pixel is a reliable way to improve representation learning for surgical scenes.

\subsection{Ablations and Sweeps}
\label{sec:results:sweeps}
Our main modification to DINOv2 is regressing depth from masked iBOT patches during pre-training. We therefore perform experiments to sweep different $\lambda_{depth}$ and the depth decoder architecture. For compute efficiency, these experiments use ViT-B models trained for a reduced budget of 115k iterations at batch size 256; results are reported in Table~\ref{tab:depth_sweep}.

An MLP (3 layers, hidden dim 512, GELU) gives the best phase recognition on Cholec80 (65.5 F1), while a transformer decoder (4 layers, embed\_dim=384, num\_heads=6) gives the best segmentation on EndoVis18 (31.2 mIoU). Performance is also sensitive to $\lambda_{depth}$: too large a weight (0.25) destabilizes training, confirming 0.1 as a robust choice. We adopt the MLP for its simplicity. We report sweep experiments over different MLP hidden dims in the appendices. 

\begin{table}[t]
  \centering
  
  \setlength{\tabcolsep}{6pt}
  \small
  \begin{tabular}{llcc}
    \toprule
    Depth decoder & $d_w$ & EV18 (mIoU) & Cholec80 (F1) \\
    \midrule
    --     & 0     & 29.3 & 62.4 \\
    \midrule
    \multirow{5}{*}{MLP}
           & 0.025 & 29.2 & 62.2 \\
           & 0.05  & 29.6 & 64.0 \\
           & 0.1   & 30.6 & \textbf{65.5} \\
           & 0.25  &  7.8 & 17.2 \\
    \midrule
    \multirow{3}{*}{Transformer}
           & 0.05  & 18.5 & 37.2 \\
           & 0.1   & \textbf{31.2} & 64.3 \\
           & 0.15  & 26.7 & 56.4 \\
    \bottomrule
  \end{tabular}
  \caption{\textbf{Depth decoder and loss-weight sweep.} EndoVis18 (EV18) linear (lin.)
  mIoU as a function of the depth decoder architecture and the depth-loss weight
  $d_w$. The top row ($d_w=0$) is the depth-free baseline; the Transformer decoder
  follows the MAE decoder design.}
  \label{tab:depth_sweep}
\end{table}

\subsection{PCA Visualization}

\begin{figure}
    \centering
    \includegraphics[width=0.8\linewidth]{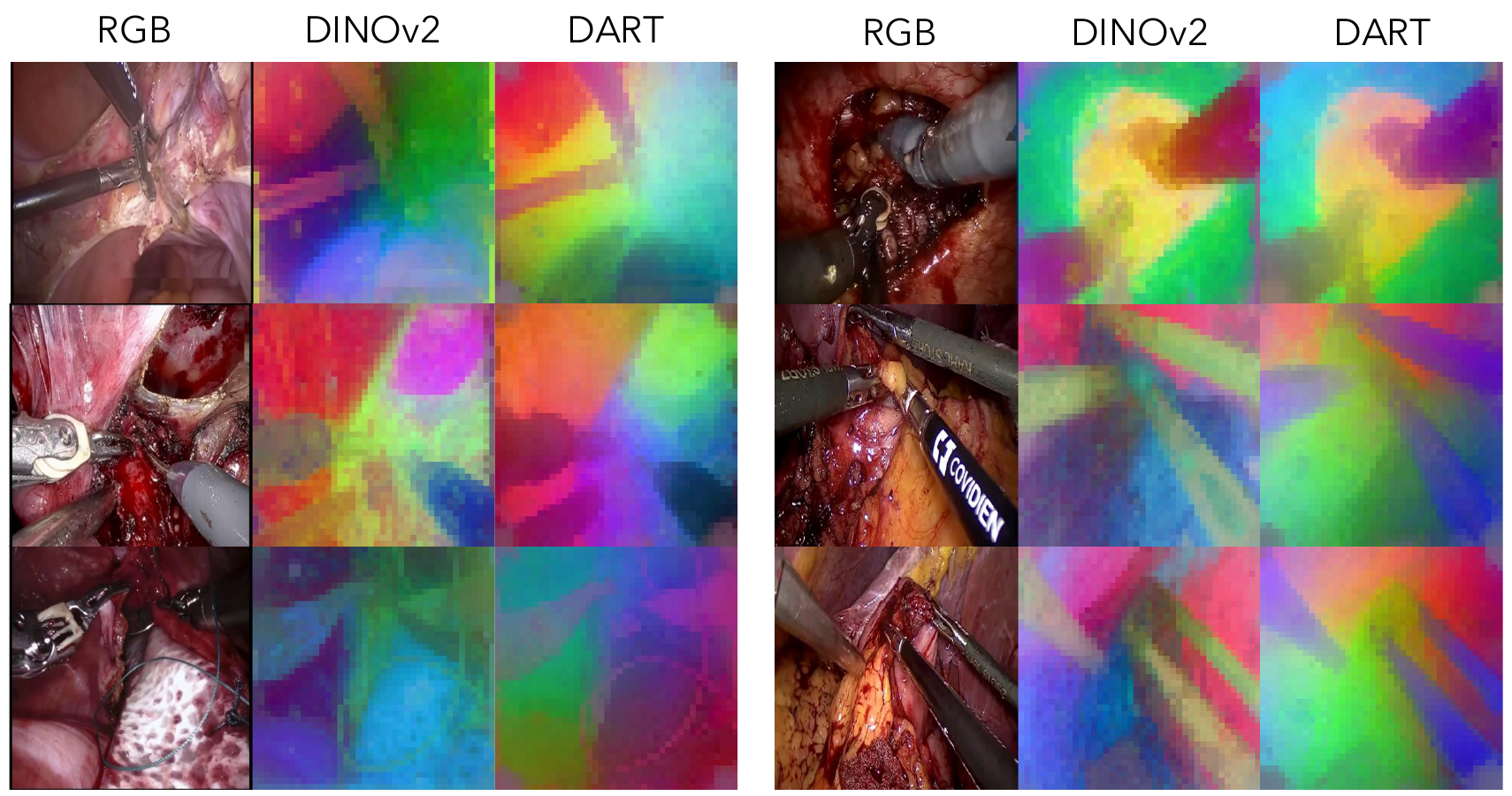}
    \caption{PCA visualization of the features learned by DINOv2 compared to \mymodel{} at $640\times640$ resolution. Generally, DART has smoother features which is reflected in its superior quantitative results. Because we do not incorporate registers in our ViTs, we clip outliers outside of 95\% for improved visualization.}
    \label{fig:pca_vis}
\end{figure}

To probe what DART's features look like qualitatively, we visualize the first three principal components of the patch features (projected to RGB) for \mymodel{} and the vanilla DINOv2 baseline trained on identical data in Fig.~\ref{fig:pca_vis}. Two differences are visible. First, DART's projections are spatially more coherent within object regions: instruments and tissue each appear as contiguous color blobs, whereas the baseline's projections show finer-grained speckle within what should be a single region. Second, boundaries between instruments, tissue, and background are crisper in \mymodel{}, with higher robustness to specularities and bright spots. Both observations are consistent with the depth-reconstruction objective: requiring the encoder to predict pseudo-depth at masked patch positions encourages features that carry both region-level consistency and geometric discontinuity at boundaries.

\section{Conclusion and Future Work}

In this work, we propose \mymodel{}, an RGB-D pretraining recipe for surgical scenes. We find that reconstructing pixel-wise depth at masked iBOT positions within the DINOv2 recipe improves performance on both dense pixel and image-level tasks. Notably, this holds even though DINOv2 otherwise operates entirely in latent space, suggesting that a targeted pixel-space objective can complement rather than disrupt self-distillation. Our auxiliary-target ablation further shows that depth provides useful regularization in ways that other spatial maps, such as Canny edges and blurred grayscale, do not. Across a wide range of tasks spanning segmentation, depth estimation, classification, and recognition, \mymodel{} delivers consistent gains over comparable baselines. Our method only uses RGB at fine-tuning and inference, improving our model's practicality for deployment.

Our method also has limitations. Its supervision relies on pseudo-labeled depth from off-the-shelf monocular estimators, which output imperfect depth. We do not characterize how sensitive the learned representations are to depth quality or the choice of the estimator. Nonetheless, there are still gains despite the noisy depth targets. Our evidence is also confined to surgical data and to ViT-\{S,B\} backbones, so whether \mymodel{} is effective at larger scales or in other domains remains an open question.

This motivates several directions for future work. Scaling to larger backbones and more data is a natural next step, as is studying robustness to the depth source. We also plan to incorporate further modalities during pretraining for richer scene understanding, such as semantic maps or robot kinematics. More broadly, our findings suggest that combining latent-space objectives (DINO, I-JEPA) with pixel-space reconstruction (MAE) is a promising direction for self-supervised pretraining, especially in specialized domains where publicly available backbones do not generalize well.

\bibliography{egbib}

\clearpage
\setcounter{page}{1}
\renewcommand{\thepage}{S\arabic{page}}   

\setcounter{section}{0}
\setcounter{figure}{0}
\setcounter{table}{0}
\setcounter{equation}{0}
\renewcommand{\thesection}{S\arabic{section}}
\renewcommand{\thefigure}{S\arabic{figure}}
\renewcommand{\thetable}{S\arabic{table}}
\renewcommand{\theequation}{S\arabic{equation}}

\begin{center}
{\Large\bf Supplementary Material}\\[2pt]
{\large DART: Depth-as-Target Pretraining for Surgical Vision Foundation Models}
\end{center}
\vspace{1em}

\section{Implementation Details}
In this section, we describe the implementation details of pre-training and fine-tuning the models used in our work. Due to computational constraints, we could not match the scale of prior work, e.g., training MAE for 1,600 epochs. However, our ``undertrained'' models outperform publicly available off-the-shelf models due to the domain gap between surgical and natural images. Pretraining implementation details are outlined in Table~\ref{tab:pretrain_config} and finetuning details are outlined in Table~\ref{tab:finetune_config}.

\begin{table}[t]
\centering
\footnotesize
\setlength{\tabcolsep}{3pt}
\renewcommand{\arraystretch}{1.1}
\caption{Pre-training configurations for in-domain baselines and DART, all trained on the LEMON corpus. ``---'' denotes parameters not applicable to a given method.}
\label{tab:pretrain_config}
\begin{tabular}{l cccc}
\toprule
\textbf{Config} & \textbf{MAE} & \textbf{MultiMAE}$^{\dagger}$ & \textbf{DINOv2} & \textbf{DART} \\
\midrule
\multicolumn{5}{l}{\textit{Optimization}} \\
Optimizer        & \multicolumn{4}{c}{AdamW} \\
Base LR          & $1\mathrm{e}{-4}$ & $1\mathrm{e}{-4}$ & $1.5\mathrm{e}{-3}$ & $1.5\mathrm{e}{-3}$ \\
Weight decay     & 0.05 & 0.05 & 0.04 & 0.04 \\
$\beta_1,\beta_2$ & 0.9, 0.95 & 0.9, 0.95 & 0.9, 0.999 & 0.9, 0.999 \\
Batch size       & 4096 & 2048 & 1024 & 1024 \\
\midrule
\multicolumn{5}{l}{\textit{Schedule}} \\
LR schedule      & \multicolumn{4}{c}{Cosine decay} \\
Warmup           & 40 ep. & 20 ep. & 62.5k it. & 62.5k it. \\
Duration         & 400 ep. & 400 ep. & 187.5k it. & 187.5k it. \\
Augmentations    & RRC & RRC & DINOv2$^{\ddagger}$ & DINOv2$^{\ddagger}$ \\
\midrule
\multicolumn{5}{l}{\textit{DINOv2-specific}} \\
Prototypes              & --- & --- & 65k & 65k \\
Crop size (G / L)       & --- & --- & 224 / 96 & 224 / 96 \\
Num.\ crops (G / L)     & --- & --- & 2 / 8 & 2 / 8 \\
Teacher momentum        & --- & --- & $0.996 \!\to\! 1$ & $0.996 \!\to\! 1$ \\
\midrule
\multicolumn{5}{l}{\textit{Multimodal supervision}} \\
Modalities                  & RGB & RGB + D & RGB & RGB + D \\
$\lambda_{\text{depth}}$    & --- & --- & --- & 0.1 \\
\bottomrule
\end{tabular}

\vspace{4pt}
{\scriptsize
$^{\dagger}$ Pre-trained with RGB and depth; fine-tuned with RGB only unless otherwise specified.\par
$^{\ddagger}$ RandomResizedCrop (RRC) plus DINOv2's standard augmentation stack.\par}
\end{table}

\begin{table}[t]
\centering
\footnotesize
\setlength{\tabcolsep}{6pt}
\renewcommand{\arraystretch}{1.15}
\caption{Fine-tuning configuration used across downstream segmentation experiments. The backbone is frozen and only the task head is trained, so depth supervision affects evaluation solely through the learned representations. Input resolution is $224\times224$ with patch size $16$. We normalize images based on the pre-training data source, e.g., ImageNet-pretrained models undergoes ImageNet normalization.}
\label{tab:finetune_config}
\begin{tabular}{lccc}
\toprule
\textbf{Config} & \textbf{lin.\ + ms.} & \textbf{DPT} & \textbf{Image-level Tasks} \\
\midrule
\multicolumn{4}{l}{\textit{Optimization}} \\
Optimizer              & AdamW         & AdamW & SGD \\
Learning rate          & $1\mathrm{e}{-3}$ & 1e-3 & 1e-3 \\
$\beta_1, \beta_2$     & 0.9, 0.999    & 0.9, 0.999 & -- \\
Weight decay           & 0.05          & 0.05 & 0.0 \\
Batch size             & 64            & 64 & 256 \\
\midrule
\multicolumn{4}{l}{\textit{Schedule}} \\
Training duration      & 50 epochs     & 100 epochs & 20 epochs \\
LR schedule            & Constant      & Constant &  Cosine Decay \\
\midrule
\multicolumn{4}{l}{\textit{Augmentations (training)}} \\
Random horizontal flip & $p = 0.5$     & $p = 0.5$ & $p=0.5$\\
Photometric distortion & Color jitter  & Color jitter & None \\
\bottomrule
\end{tabular}
\end{table}

\begin{table}[t]
\centering
\footnotesize
\setlength{\tabcolsep}{3pt}
\renewcommand{\arraystretch}{1.2}
\caption{Downstream evaluation datasets used in this work. ``In/Ex'' denotes \emph{in-vivo} versus \emph{ex-vivo}. ``Classes'' refers to the number of label categories used in this work. Lap.\ = laparoscopy.}
\label{tab:downstream_datasets}
\begin{tabular}{l l l l c c l}
\toprule
\textbf{Dataset} & \textbf{Task} & \textbf{Procedure} & \textbf{Modality} & \textbf{In/Ex} & \textbf{Cls.} & \textbf{Size} \\
\midrule
Cholec80           & Phase recog.\        & Cholecystectomy    & Manual lap.\ & In & 7   & 80 videos \\
CholecT50          & Triplet recog.\      & Cholecystectomy    & Manual lap.\ & In & 100$^{\ast}$ & 50 vid.\ / ${\sim}100$k fr.\ \\
SCARED-C           & Depth estimation     & Porcine abdomen    & Robotic      & Ex & ---  & 17{,}135 RGB-D fr.\ \\
EndoVis18          & Scene seg.\          & Porcine nephrectomy & Robotic     & Ex & 12   & 19 video seq.\ \\
CholecInstanceSeg  & Instrument seg.\     & Cholecystectomy    & Manual lap.\ & In & 8    & 41{,}933 fr.\ / 85 proc.\ \\
SAR-RARP50         & Instrument seg.\     & Radical prostatectomy & Robotic   & In & 10   & 50 vid.\ / ${\sim}16$k fr.\ \\
CholecSeg8k        & Scene seg.\          & Cholecystectomy    & Manual lap.\ & In & 13   & 8{,}080 fr.\ / 17 clips \\
PhaKIR             & Instrument seg.\     & Cholecystectomy    & Manual lap.\ & In & 20  & 19{,}435 fr.\ \\
\bottomrule
\end{tabular}

\vspace{2pt}
{\scriptsize $^{\ast}$ 100 valid action triplets formed from 6 instruments, 10 verbs, and 15 targets.}
\end{table}

\subsection{LEMON and RGB-D Dataset Generation}
The LEMON dataset is a surgical video dataset comprised of 938 hours curated from YouTube. We sample frames at 1\,Hz from every video, which results in 3.4M images. We use the ViT-L variant of Depth Anything V2 to extract relative depth maps for each image. These RGB-D pairs are compressed into shards for efficient training with the \verb|webdataset| library. Because Depth Anything is a monocular model, the estimated depth maps are natively inverse depth, or disparity. The authors do this to account for different scales between datasets, e.g., indoor and outdoor. As can be seen in our qualitative results, we maintain inverse depth during pre-training; higher values are closer to the camera. Although we did not experiment with inverting to relative depth, we do not believe it would lead to significant results since our framework does not require physically meaningful depth values during training.

This approach is not without limitations. For instance, the LEMON dataset contains frames with black borders, common in endoscopic video feed. Although the depth values should be invalid in these zones, we use the raw depth outputs in these border areas. This may be included in future work. Furthermore, monocular depth estimation is inherently ill-posed; we saw that a minority of depth samples were inaccurate due to unclean images, blurring, or bloody scenes. Nonetheless, we see that noisy depth maps are still a valuable source for improving representation quality.

\subsection{Downstream Task Datasets}
We use 8 labeled datasets for downstream tasks in our work. We give some dataset details in Table~\ref{tab:downstream_datasets}.

\section{ViT-B vs ViT-S Experiments}
We also provide results for ViT-S. However, due to computational constraints, we train these with batch size 256 rather than 1024, which explains the worse results of ViT-B in this table compared to the main results. Similar to previous work, we noticed that batch size was a major contributing factor to good performance of DINOv2, since the centering mechanism and Koleo regularization depends on a large minibatch for good statistics. ViT-S vs. ViT-B results are reported in Table~\ref{tab:vits_vitb}.

\begin{table}
  \centering
  \setlength{\tabcolsep}{2.5pt}
  \footnotesize
  \begin{tabular}{@{}ll c cc cc cc cc cc@{}}
    \toprule
    & & & \multicolumn{2}{c}{EndoVis18} & \multicolumn{2}{c}{CholecISeg} & \multicolumn{2}{c}{SARRARP50} & \multicolumn{2}{c}{CholecSeg8k} & \multicolumn{2}{c}{PhaKIR} \\
    \cmidrule(lr){4-5} \cmidrule(lr){6-7} \cmidrule(lr){8-9} \cmidrule(lr){10-11} \cmidrule(lr){12-13}
    Data & Method & Depth & lin. & +ms. & lin. & +ms. & lin. & +ms. & lin. & +ms. & lin. & +ms. \\
    \midrule
    \multirow{4}{*}{LEMON}
    & DINOv2 ViT-S          & --    & 30.4 & 30.9 & 48.5 & \bf 50.9 & 57.4 & 54.6 & 39.9 & 42.7 & 10.3 &  9.5 \\
    & \textbf{Ours ViT-S}   & PT    & \bf 31.4 & \bf 31.2 & \bf 48.5 & 47.4 & \bf 57.8 & \bf 55.6 & \bf 41.0 & \bf 42.7 & \bf 10.7 & \bf 10.7 \\
    \cmidrule(lr){2-13}
    & DINOv2 ViT-B          & --    & 32.1 & 31.8 & 52.8 & 56.3 & 61.5 & 65.1 & 41.6 & 46.1 & 11.8 & 12.8 \\
    & \textbf{Ours ViT-B}   & PT    & 32.1 & \bf 32.2 & \bf 57.2 & \bf 63.5 & \bf 63.5 & \bf 67.0 & \bf 43.9 & \bf 50.3 & \bf 12.3 & \bf 13.6 \\
    \bottomrule
  \end{tabular}
  \caption{\textbf{ViT-S semantic segmentation with frozen features}, linear (lin.) and
  multiscale (+ms), with ViT-B counterparts for reference. \emph{Depth} marks whether
  depth is used in pre-training (PT) and/or fine-tuning (FT). Mean over 3 seeds.
  CholecISeg = CholecInstanceSeg. For these experiments, we use batch size = 256 due to compute constraints.}
  \label{tab:vits_vitb}
\end{table}

\section{ViT-S Depth Decoder Sweeps}
We noticed in our experiments that the ViT-S models did not perform expectedly with the same depth decoder size as the ViT-B models. As a result, we perform a small sweep of smaller MLP decoder to get the best ViT-S results for DART. Specifically, we sweep over hidden dim sizes $\{128, 256, 512\}$, shown in Table~\ref{tab:vits_sweep}.

\begin{table}[t]
\centering
\footnotesize
\setlength{\tabcolsep}{4pt}
\renewcommand{\arraystretch}{1.15}
\caption{Depth-head hidden dimension sweep (ViT-S). Linear (lin.) and multiscale (+ms.) frozen-feature mIoU on four surgical segmentation benchmarks. Hidden dimensions are reported in ascending order; the leftmost column is the depth-free DINOv2 baseline. Best per row in \textbf{bold}.}
\label{tab:vits_sweep}
\begin{tabular}{l l cccc}
\toprule
\textbf{Dataset} & \textbf{Head} & \textbf{DINOv2} & \textbf{DART-128} & \textbf{DART-256} & \textbf{DART-512} \\
\midrule
\multirow{2}{*}{EndoVis18}        & lin.  & 30.4 & 29.6 & \textbf{31.4} & 29.5 \\
                                  & +ms.  & 30.9 & \textbf{31.4} & 31.2 & 30.8 \\
\midrule
\multirow{2}{*}{CholecInstanceSeg} & lin.  & \textbf{48.5} & 47.2 & \textbf{48.5} & 46.8 \\
                                  & +ms.  & \textbf{50.9} & 47.2 & 47.4 & 46.4 \\
\midrule
\multirow{2}{*}{CholecSeg8k}      & lin.  & 39.9 & \textbf{42.9} & 41.0 & 39.1 \\
                                  & +ms.  & \textbf{42.7} & 42.5 & \textbf{42.7} & 40.3 \\
\midrule
\multirow{2}{*}{PhaKIR}           & lin.  & 10.3 & \textbf{11.6} & 10.7 & 10.0 \\
                                  & +ms.  & 9.5  & 10.5 & \textbf{10.7} & 10.4 \\
\bottomrule
\end{tabular}
\end{table}

\section{Full Segmentation Results}
In this section, we provide the standard deviation values in addition to the mean of the segmentation results from the main results. For all runs, we ran with 3 different seeds which we've reported the mean of. We could not include the standard deviation due to space constraints, so we provide them here in Table~\ref{tab:semseg-std-1} and Table~\ref{tab:semseg-std-2}.

\begin{table}
  \centering
  \setlength{\tabcolsep}{2.5pt}
  \footnotesize
  \newcommand{\dimrow}[1]{\textcolor{gray}{#1}}
  \newcommand{\sd}[1]{{\tiny$\pm$#1}}
  \begin{tabular}{@{}ll c cc cc cc@{}}
    \toprule
    & & & \multicolumn{2}{c}{EndoVis18} & \multicolumn{2}{c}{CholecInstanceSeg} & \multicolumn{2}{c}{SARRARP50} \\
    \cmidrule(lr){4-5} \cmidrule(lr){6-7} \cmidrule(lr){8-9}
    Data & Method & Depth & lin. & +ms. & lin. & +ms. & lin. & +ms. \\
    \midrule
    \multirow{4}{*}{Natural}
    & DINOv2   & --    & 28.8\,\sd{1.0} & 31.8\,\sd{0.7} & 48.3\,\sd{0.2} & 52.6\,\sd{0.5} & 53.5\,\sd{0.4} & 62.0\,\sd{0.2} \\
    & DINOv3   & --    & 30.1\,\sd{0.5} & 32.5\,\sd{0.6} & 53.5\,\sd{2.7} & 55.2\,\sd{4.1} & 58.0\,\sd{0.8} & 64.0\,\sd{0.4} \\
    & MAE      & --    & 26.0\,\sd{0.2} & 28.0\,\sd{0.4} & 37.1\,\sd{0.3} & 47.1\,\sd{0.4} & 52.7\,\sd{0.2} & 55.1\,\sd{0.5} \\
    & MultiMAE & PT    & 26.6\,\sd{0.3} & 27.0\,\sd{0.5} & 40.7\,\sd{0.3} & 48.7\,\sd{0.4} & 51.2\,\sd{0.2} & 56.1\,\sd{0.1} \\
    \midrule
    \multirow{6}{*}{LEMON}
    & LEMON-FM      & --    & 25.7\,\sd{0.4} & 31.2\,\sd{0.3} & 40.6\,\sd{0.4} & 51.0\,\sd{0.1} & 52.0\,\sd{0.2} & 66.6\,\sd{0.3} \\
    & MAE           & --    & 31.9\,\sd{0.5} & 32.6\,\sd{0.4} & 41.1\,\sd{0.3} & 56.2\,\sd{2.8} & 65.7\,\sd{0.5} & 67.1\,\sd{0.1} \\
    & MultiMAE      & PT    & 30.8\,\sd{0.2} & 32.6\,\sd{0.4} & 59.5\,\sd{0.4} & 55.1\,\sd{0.3} & 62.7\,\sd{0.8} & 66.0\,\sd{0.2} \\
    & DINOv2        & --    & 34.1\,\sd{0.3} & 34.9\,\sd{0.3} & 58.2\,\sd{0.6} & 59.7\,\sd{3.4} & 63.9\,\sd{0.5} & 64.4\,\sd{0.2} \\
    & \textbf{Ours} & PT    & \textbf{34.9}\,\sd{0.3} & \textbf{36.8}\,\sd{0.4} & \textbf{66.2}\,\sd{3.7} & \textbf{68.9}\,\sd{0.3} & \textbf{68.7}\,\sd{0.4} & \textbf{70.2}\,\sd{0.8} \\
    & \dimrow{MultiMAE} & \dimrow{PT+FT} & \dimrow{29.4\,\sd{0.6}} & \dimrow{28.8\,\sd{1.2}} & \dimrow{54.7\,\sd{0.5}} & \dimrow{49.9\,\sd{0.4}} & \dimrow{70.4\,\sd{0.6}} & \dimrow{71.8\,\sd{0.2}} \\
    \bottomrule
  \end{tabular}
  \caption{\textbf{Semantic segmentation with frozen features, mean\,$\pm$\,std (part 1 of 2).}
  Linear (lin.) and multiscale (+ms.). \emph{Depth} marks whether depth is used in
  pre-training (PT) and/or fine-tuning (FT). Mean and standard deviation over 3 seeds.
  CholecISeg = CholecInstanceSeg. Natural images refer to LVD for DINO models and IN-1k
  for MAE models. Remaining datasets in Table~\ref{tab:semseg-std-2}.}
  \label{tab:semseg-std-1}
\end{table}

\begin{table}
  \centering
  \setlength{\tabcolsep}{2.5pt}
  \footnotesize
  \newcommand{\dimrow}[1]{\textcolor{gray}{#1}}
  \newcommand{\sd}[1]{{\tiny$\pm$#1}}
  \begin{tabular}{@{}ll c cc cc@{}}
    \toprule
    & & & \multicolumn{2}{c}{CholecSeg8k} & \multicolumn{2}{c}{PhaKIR} \\
    \cmidrule(lr){4-5} \cmidrule(lr){6-7}
    Data & Method & Depth & lin. & +ms. & lin. & +ms. \\
    \midrule
    \multirow{4}{*}{Natural}
    & DINOv2   & --    & 36.7\,\sd{0.5} & 38.5\,\sd{0.5} & 10.3\,\sd{0.6} & 10.4\,\sd{0.3} \\
    & DINOv3   & --    & 50.0\,\sd{2.9} & 43.7\,\sd{0.4} & 11.6\,\sd{0.3} & 10.6\,\sd{0.1} \\
    & MAE      & --    & 37.9\,\sd{2.1} & 38.9\,\sd{1.2} &  7.6\,\sd{0.8} &  8.8\,\sd{0.4} \\
    & MultiMAE & PT    & 36.0\,\sd{1.1} & 36.7\,\sd{0.4} &  8.3\,\sd{0.5} & 10.1\,\sd{0.3} \\
    \midrule
    \multirow{6}{*}{LEMON}
    & LEMON-FM      & --    & 33.9\,\sd{0.3} & 42.4\,\sd{0.8} &  8.7\,\sd{0.5} & 11.0\,\sd{0.2} \\
    & MAE           & --    & 44.3\,\sd{0.1} & 46.0\,\sd{2.2} & 12.0\,\sd{0.2} & 13.2\,\sd{0.4} \\
    & MultiMAE      & PT    & 43.4\,\sd{0.3} & 43.1\,\sd{0.2} & 11.7\,\sd{0.3} & 12.2\,\sd{0.1} \\
    & DINOv2        & --    & 49.7\,\sd{2.1} & 50.5\,\sd{2.0} & 14.1\,\sd{0.4} & 15.7\,\sd{0.4} \\
    & \textbf{Ours} & PT    & \textbf{52.4}\,\sd{0.2} & \textbf{50.5}\,\sd{1.9} & \textbf{14.5}\,\sd{0.5} & \textbf{16.2}\,\sd{0.9} \\
    & \dimrow{MultiMAE} & \dimrow{PT+FT} & \dimrow{44.3\,\sd{0.8}} & \dimrow{45.2\,\sd{0.5}} & \dimrow{11.0\,\sd{0.6}} & \dimrow{11.9\,\sd{0.8}} \\
    \bottomrule
  \end{tabular}
  \caption{\textbf{Semantic segmentation with frozen features, mean\,$\pm$\,std (part 2 of 2).}
  Continuation of Table~\ref{tab:semseg-std-1} for the CholecSeg8k and PhaKIR datasets.
  Mean and standard deviation over 3 seeds. Column conventions as in
  Table~\ref{tab:semseg-std-1}.}
  \label{tab:semseg-std-2}
\end{table}

\section{Depth Anything v2 Comparison Study}

Intuitively, a foundational depth model initialized with DINOv2 weights and subsequently trained on depth data should be competitive in dense pixel tasks. To empirically validate this, we take the ViT-B variant of DAv2, discard the depth head, freeze the backbone, and train identical heads for the segmentation and depth estimation tasks. We also maintain the same training configuration and parameters as our original paper. Results are shown in Table~\ref{tab:dav2_results}.

\begin{table}[h]
\centering
\setlength{\tabcolsep}{4pt}
\begin{tabular}{lcccc}
\toprule
 & \multicolumn{2}{c}{DAv2} & \multicolumn{2}{c}{DART} \\
\cmidrule(lr){2-3} \cmidrule(lr){4-5}
Dataset & H1 & H2 & H1 & H2 \\
\midrule
EndoVis18   & 27.1  & 33.7  & \textbf{34.9}  & \textbf{36.8}  \\
CholecISeg  & 46.5  & 51.0  & \textbf{66.2}  & \textbf{68.9}  \\
SARRARP50   & 52.0  & 59.6  & \textbf{68.7}  & \textbf{70.2}  \\
CholecSeg8k & 36.0  & 36.8  & \textbf{52.4}  & \textbf{50.5}  \\
PhaKIR      & 9.6   & 10.7  & \textbf{14.5}  & \textbf{16.2}  \\
\midrule
SCARED-C    & 0.547 & 0.612 & \textbf{0.557} & \textbf{0.653} \\
\bottomrule
\end{tabular}
\caption{Quantitative results for frozen Depth Anything v2 ViT-B on dense pixel tasks. H1 is a single linear head; H2 is +m.s. for segmentation and DPT for depth. The first five rows are segmentation (mIoU) and the last is depth estimation ($\delta_1$).}
\label{tab:dav2_results}
\end{table}

Please refer to Table 1 and 2 in the main paper to compare these values. DAv2 performs competitively on segmentation tasks and is a plausible backbone for dense pixel tasks, demonstrating superior performance compared to MAE and MultiMAE trained on ImageNet-1k. However, off-the-shelf DINOv2 and DINOv3 models slightly outperform DAv2, while the DINOv2 and DART models trained on surgical data far outperform DAv2. For depth estimation (SCARED-C), DAv2 is much more competitive, even outperforming DINOv2 trained on the surgical data. However, DART still demonstrates superior performance over DAv2. These results suggest that although DAv2 can be a useful backbone for downstream visual tasks, training a vision backbone from scratch with depth-as-target is still a better recipe for improved visual understanding. Feature-level alignment within intermediate ViT layers (e.g., \textbf{RePA} or \textbf{GeometryForcing}) rather than depth-as-target may also produce similar gains in performance; we save this for future work on comprehensive comparison studies between training regimes.

\section{Additional Qualitative Results}

We show additional qualitative results for segmentation and depth estimation tasks in this section. We include all 9 models that were evaluated in the work.

In depth estimation, we observe that some areas of the image have visually inaccurate results. Please note that the SCARED-C dataset only provides sparse depth; the lack of dense supervision limits the performance ceiling of these models.

\begin{figure}
    \centering
    \includegraphics[width=0.8\linewidth]{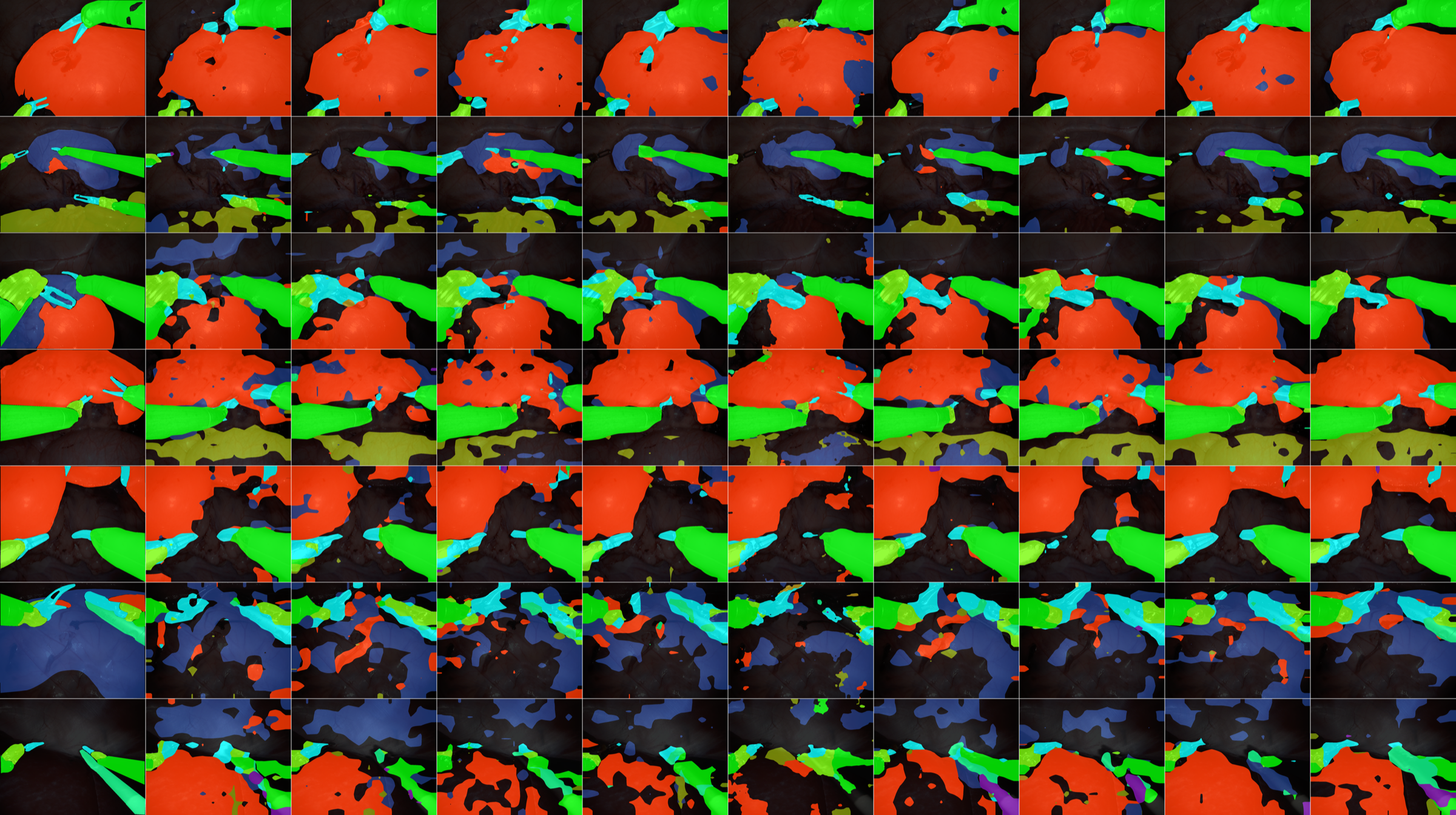}
    \caption{EndoVis18 segmentation qualitative results. From left to right: Ground truth, MAE (IN-1k), MultiMAE (IN-1k), DINOv2 (LVD), DINOv3 (LVD), LEMON-FM, MAE (LEMON), MultiMAE (LEMON), DINOv2 (LEMON), DART (LEMON).}
    \label{fig:supp_endovis18_qual}
\end{figure}

\begin{figure}
    \centering
    \includegraphics[width=0.8\linewidth]{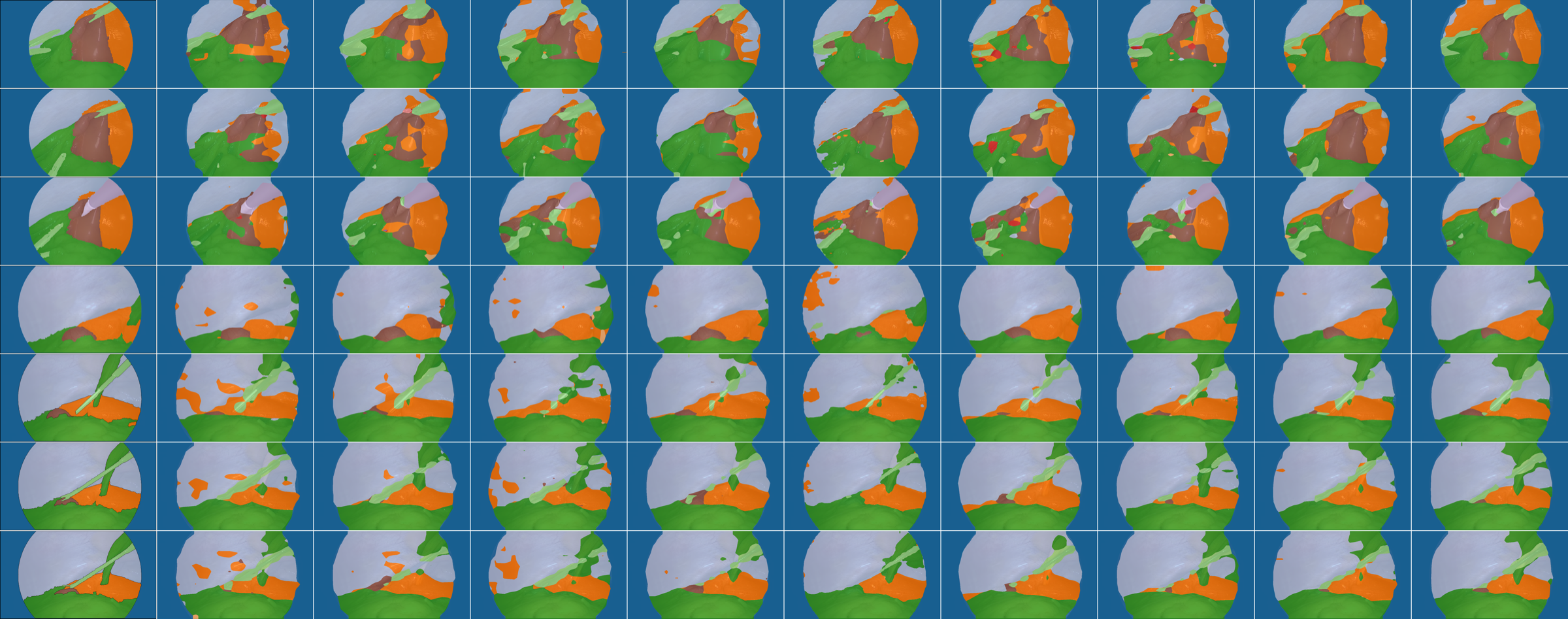}
    \caption{CholecSeg8k segmentation qualitative results. From left to right: Ground truth, MAE (IN-1k), MultiMAE (IN-1k), DINOv2 (LVD), DINOv3 (LVD), LEMON-FM, MAE (LEMON), MultiMAE (LEMON), DINOv2 (LEMON), DART (LEMON).}
    \label{fig:supp_cholecseg8k_qual}
\end{figure}

\begin{figure}
    \centering
    \includegraphics[width=0.8\linewidth]{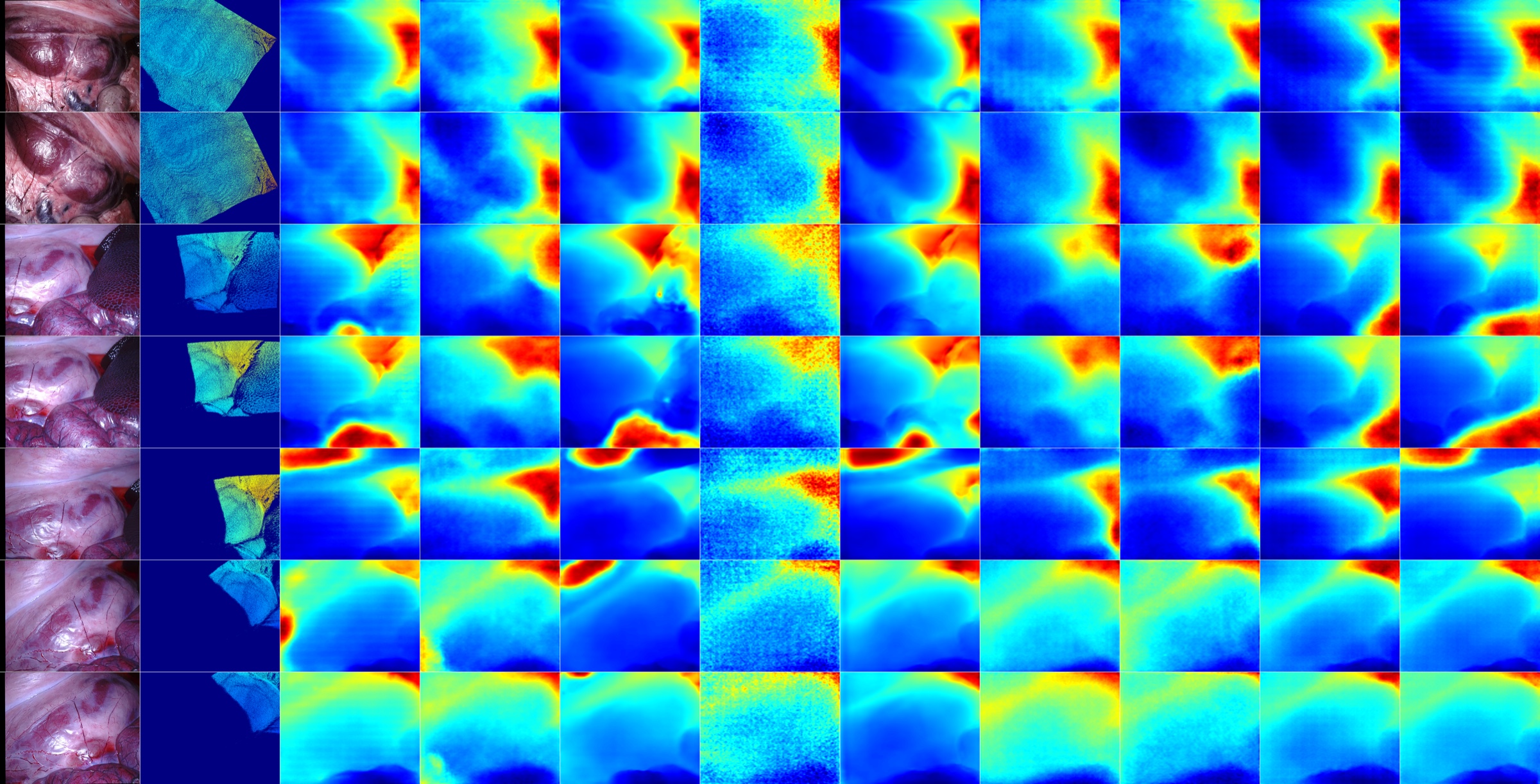}
    \caption{SCARED-C depth estimation qualitative results. From left to right: RGB, Ground truth, MAE (IN-1k), MultiMAE (IN-1k), DINOv2 (LVD), DINOv3 (LVD), LEMON-FM, MAE (LEMON), MultiMAE (LEMON), DINOv2 (LEMON), DART (LEMON).}
    \label{fig:supp_scared_qual}
\end{figure}

\end{document}